\documentclass{article}

\usepackage[preprint]{neurips_2021}

\usepackage[utf8]{inputenc} 
\usepackage[T1]{fontenc}    
\usepackage{hyperref}       
\usepackage{url}            
\usepackage{booktabs}       
\usepackage{amsfonts}       
\usepackage{nicefrac}       
\usepackage{microtype}      
\usepackage{xcolor}         

\usepackage{microtype}
\usepackage{graphicx}
\usepackage{subfigure}
\usepackage{amsmath}
\usepackage{wrapfig}
\usepackage{enumitem}
\usepackage{amsfonts}
\usepackage{booktabs} 

\usepackage{amsmath}
\usepackage{algorithm,algpseudocode}

\usepackage{hyperref}

\title{\Large Molecular Attributes Transfer from Non-Parallel Data}

\author{%
     {Shuangjia Zheng}\thanks{Both authors contribute equally to the work.}  \\
  Sun Yat-sen University, Galixir\\
  zhengshj9@mail2.sysu.edu.cn\\
  \And
 Ying Song\footnotemark[1]\\
  Sun Yat-sen University\\
 songy75@mail2.sysu.edu.cn 
   \And
    Pan Zhang \\
    Sun Yat-sen University\\
    zhangp258@mail2.sysu.edu.cn\\

  \And
    Chengtao Li  \\
    Galixir \\
    chengtao.li@galixir.com \\
  \And
    Le Song \\
    Mohamed bin Zayed University of AI \\
    le.song@mbzuai.ac.ae
 \\
  \And
    Yuedong Yang \\
    Sun Yat-sen University \\
    yangyd25@mail.sysu.edu.cn \\
 
}

\begin{document}

\maketitle

\begin{abstract}

Optimizing chemical molecules for desired properties lies at the core of drug development. Despite initial successes made by deep generative models and reinforcement learning methods, these methods were mostly limited by the requirement of predefined attribute functions or parallel data with manually pre-compiled pairs of original and optimized molecules. In this paper, for the first time, we formulate molecular optimization as a style transfer problem and present a novel generative model that could automatically learn internal differences between two groups of non-parallel data through adversarial training strategies. Our model further enables both preservation of molecular contents and optimization of molecular properties through combining auxiliary guided-variational autoencoders and generative flow techniques. Experiments on two molecular optimization tasks, toxicity modification and synthesizability improvement, demonstrate that our model significantly outperforms several state-of-the-art methods.

\end{abstract}

\section{Introduction}

\setlength{\abovedisplayskip}{3pt}
\setlength{\abovedisplayshortskip}{3pt}
\setlength{\belowdisplayskip}{3pt}
\setlength{\belowdisplayshortskip}{3pt}
\setlength{\jot}{2pt}
\setlength{\floatsep}{1ex}
\setlength{\textfloatsep}{1ex}
\setlength{\intextsep}{1ex}
\setlength{\parskip}{1ex}

Optimizing chemical molecules for desired properties lies at the core of drug development. During the optimization process, the key objective is to maintain favorable attributes (e.g., physicochemical properties, potency) in reference compounds while improving on deficiencies (e.g., toxicity, synthetic accessibility) within the reference structure\cite{nicolaou2007molecular}. It is challenging to meet these criteria simultaneously,  partly due to the immense size of the chemical space. The range of synthesizable small molecules has been estimated to be between $10^{23}$ and $10^{60}$ \cite{polishchuk2013estimation}, making exhaustive search hopeless. On the other hand, the lack of paired molecules that conform to all the constraints hinders the development of reliable supervised learning models. 

Over the past few years, there have been significant advances in applying deep learning models to molecule optimization (MO). Existing solutions can be divided into two main categories, either utilizing an unbiased generative model to generate abundant candidates and then performing \emph{post hoc} filtering or learning a continuous vector representation of the discrete molecular structures and optimizing with a reward function composed of desired molecular property objectives \cite{blaschke2020reinvent,you2018graph}. While these methods have made initial successes, they either are computationally expensive or perform conditional generation obliquely. In addition, most of the works only optimize a single objective while ignoring the preservation of the original input characteristics, limiting their usage in real-world molecular optimization paradigm where certain physio-chemical properties need to be reserved in order to develop viable drug candidates \cite{lin1997role}. Though multi-objective molecule optimization was investigated in a few attempts of using reinforcement learning or genetic algorithm  \cite{li2018multi,winter2019efficient,jin2020multi}, these works do not input any specific input molecule as the anchor, leading to the generation of randomized or uncontrollable molecules. 

Recently, MO was re-formulated  as a supervised graph-to-graph transformation task \cite{jin2018learning} with artificial molecular pairs in supervised settings. It takes the characteristics of the input molecule into consideration and provides good attribute transfer results. Unfortunately, this strategy is unable to apply to most of the molecular optimization scenarios where the molecular pairs do not exist. Instead, only non-parallel data is available with collections of molecules from each domain (e.g., toxic, non-toxic) without explicit correspondence. Therefore, it is necessary to formulate new ways to transfer molecular attributes using the non-parallel data of not predefined molecule pairs.

\begin{wrapfigure}{r}{0.5\textwidth} 
\vspace{-9pt}
\centering
\def\arraystretch{1}
{
    \includegraphics[width=0.5\textwidth]{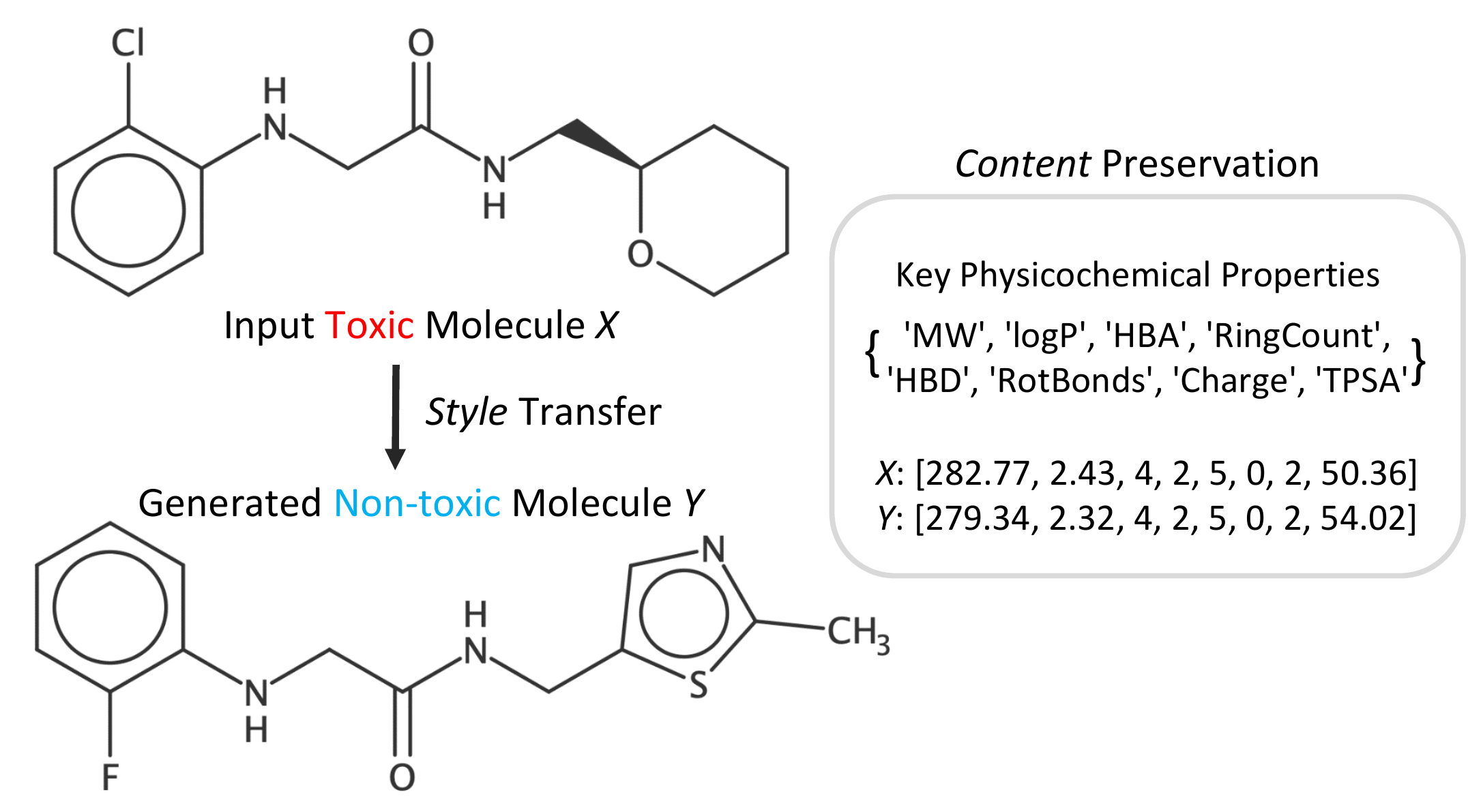}
    \caption{An example of molecular attribute transfer. The goal is to convert some attributes of a molecule (e.g., toxic) to other attributes (e.g., non-toxic) while preserving attribute-independent content. }
    
}
\label{fig:example}
\vspace{-9pt}
\end{wrapfigure}

In this paper, we tackle the problem of molecular attributes transfer, \emph{aka} molecular style transfer, which aims to endow a source molecule with different attributes (e.g. transform the toxicity from toxic to non-toxic), while maintaining the attribute-independent content as the source molecule (e.g. molecule weight, the number of ring) based on two groups of non-parallel data. To enable the manipulation of generated molecules, a few challenges need to be addressed.

The first challenge comes from the discrete nature of molecular samples \cite{you2018graph}. The resulting non-differentiable issue hinders the use of global discriminators that evaluate generated molecules and gradient propagation that guides the optimization of generators in a holistic manner. 

Another challenge for molecular attributes transfer relates to learning disentangled latent representations. Similar to text, the attribute of molecule is a broad concept and refers to any characteristic or property of molecule, which can be related to content or style. Moreover, a small modification of molecular structure may result in a series of changes in chemical properties \cite{maggiora2014molecular}. As a result, the natural entanglement of \emph{content} and \emph{style} will cause a contradiction when learning disentangled latent representations. 

In this paper, we develop a novel molecular SMILES \cite{weininger1988smiles} string-based unsupervised attribute transfer algorithm that approximate the ultimate goal of molecule optimization. Different from previous work that relied on heuristic score or parallel data with manually precompiled paired molecules, we focus on generating expected compounds whose target attributes can be controlled by editing the latent representation of input molecule in continuous space with non-parallel corpora. In particular, we first pretrain an auxiliary guided-VAE to encourage the latent representation to be chemically interpretable, ensuring that produced molecules adhere to the original content they are conditioned on. Then, to achieve a better style transfer accuracy, our model incorporates a set of molecular instances sharing the same style, and learns to extract underlying stylistic properties by forming a more discriminative style latent space with the generative flow technique. Finally, we design a novel adversarial discriminator and a systematic set of auxiliary losses to enforce the separation and fusion of the \emph{style} and \emph{content} in continuous latent space.


\section{Related work}

\noindent \textbf{Generative model for molecular optimization.} Over the past few years, there has been a surge of interest in using the generative model to discover novel functional molecules with certain properties \cite{gomez2018automatic,olivecrona2017molecular, jin2018junction, zheng2019qbmg, you2018graph}. They represented molecules as either SMILES strings or molecular graphs, and used Reinforcement Learning and Bayesian optimization to optimize the properties of the generated molecules. While most of the researches only pose a single property constraint, multiple properties need to be considered in order to develop viable drug candidates. The more recent works try substructure-based \cite{jin2020multi,lim2020scaffold} molecular evolution to optimize multiple properties that are a highly related but different task since they directly put constraints on the structure rather than attributes.

Most similar to our approach is \cite{jin2018learning} which formulates this problem as graph-to-graph translation and outperforms previous methods in the conditional setting. They used molecule structural similarity to artificially construct a parallel training set with  paired molecules, limiting the diversity of the target chemical space. Several variants \cite{fu2020core,jin2019hierarchical,zheng2020deep} have been proposed to modify the unstable training process but still remain imperfect due to the unpractical supervision settings. Compared to previous works, our approach is the first to perform unsupervised molecule attributes transfer on non-parallel datasets.


\noindent \textbf{Text style transfer.} The text style transfer task aims to change the stylistic properties (e.g., sentiment) of the text while retaining the style-independent content within the context. In particular, our work is closely related to the non-parallel text attribute transfer task \cite{shen2017style,hu2017toward}. The mainstream strategy is to formulate the style transfer problem into the “encoder-decoder” framework by explicitly disentangling the content and style in the latent space and then combining the content with a target style to achieve a transfer \cite{shen2017style,lample2018multiple,john2018disentangled}.  Most solutions adopt an adversarial paradigm to learn latent embeddings agnostic to the original style of input sentences. However, the adversarial strategies on sentence generation usually require reinforcement learning \cite{shi2018toward} or an approximation of the output softmax weights with adjustable temperatures \cite{hu2017toward} to handle the discrete issues, which tend to be slow, erratic, and difficult to control. This could be particularly serious in molecular generation \cite{sanchez2017optimizing}, since the grammar of molecular SMILES string is more delicate and the slightest perturbation can bring about syntax issues \cite{dai2018syntax}. Different from previous methods, we perform a novel adversarial regularization over continuous latent space. This allows our model to focus on sampling desired latent variables and not worry about syntax issues.




\section{Methods}

\subsection{Problem formulation}

In this work, we represent a molecule structure by its SMILES string \cite{faulon2003signature}, which is a sequence of special tokens representing atoms, branches as well as opening and closure of rings. We define two specific attributes, synthesizability and toxicity, respectively as the \emph{style} of molecules, and eight basic physical attributes (as presented in Figure \ref{fig:example}) as the \emph{content} of molecules. The generated candidates should meet the following requirements: (i) conforming to the target style, (ii) preserving the content attributes in the original input molecule, and (iii) ensuring the validity of generated molecules. 

\begin{wrapfigure}{r}{0.5\textwidth} 
\vspace{-2mm}
\centering
\def\arraystretch{1}
{
    \includegraphics[width=0.5\textwidth]{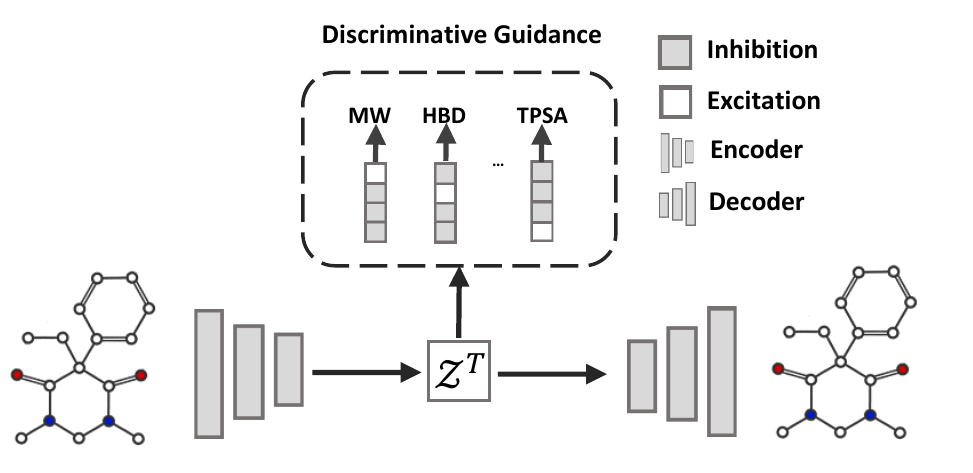}
    \vspace{-5mm}
    \caption{An example of molecular attribute transfer. The goal is to convert some attributes of a molecule (e.g., toxic) to other attributes (e.g., non-toxic) while preserving attribute-independent content. }
    \label{fig:fig2}
    
}

\vspace{-15pt}
\end{wrapfigure}
Formally, suppose there are two datasets $\mathcal{D}^i$ and $\mathcal{D}^j$, and molecules in $\mathcal{D}^i$ ($\mathcal{D}^j$) share the same style $s^i$ ($s^j$). Given an arbitrary molecule $m^i$ from $\mathcal{D}^i$ with style $s^i$, our goal is to learn a transformation $\mathcal{G}$: $m^i \rightarrow \hat{m}^j$, where $\hat{m}^j$ has a target style $s^j$ while preserving the content in original molecule $m^i$ as much as possible.



\subsection{Overview}
Our novel framework is trained on a non-parallel but style-labeled corpus. We accomplish the attribute transfer with a disentanglement process. In particular, we first pre-train an auxiliary guided autoencoder as our base model (section 3.3), containing an encoder $Enc(x; \phi)$ and a decoder $Dec(z; \theta)$. Based on this pre-trained module, we use $Enc(x; \phi)$ to encode the reference compound $x$ with desired content into $z_c$ and the molecule $m^j$ with desired style into $z_s$. A generative flow technique $GF$ is introduced to extract a comprehensive stylistic feature $h_s$ from a batch of \textit{molecular style instances} $ \Phi_{K}^{j} \subset \mathcal{D}^j$ with the same style (section 3.4). We then force the latent representation $z_c$ to contain all information except the unfavorable attribute via adversarial learning (section 3.5). The transferred output $z_g$ is computed based on the combination of $z_c$ and the stylistic attribute code $h_s$. Finally, We use $Dec(z_g; \theta)$ to reverse the latent variable to novel molecules. 




\subsection{Learning Content Latent Space}
\noindent \textbf{Variational Autoencoder.} 
Let $x = (o_1, ..., o_n)$ be an input molecular SMILES string with $n$ tokens. A variational Autoencoder (VAE) encodes a molecule $x$ to a latent vector space, from which the input is reconstructed.
Following the standard paradigm in VAE \cite{kingma2013auto}, the latent variables are denoted by vector $z$. The encoder $Enc(x; \phi)$ includes network and variational parameters $\phi$ that constructs variational probability model $q_\phi(z|x)$. The decoder network  $Dec(z; \theta)$ is parameterized by $\theta$ to reconstruct samples $\widetilde{x} = \mathcal{f}_\theta(z)$. Both the encoder and decoder employ three layers of RNNs with Gated Recurrent Unit (GRU) \cite{chung2014empirical}. The estimation of log likelihood log $p(x)$ is achieved by maximizing the evidence lower bound (ELBO) \cite{kingma2013auto}:
\begin{equation}
\mathcal{L}_{KL}=-\mathrm{{\lambda}_{kl}}{\mathcal{D}_{KL}}(q_\phi(z|x)||p(z))+\mathbb{E}_{q_\phi(z|x)}[\mathrm{log}p_{\theta}(x|z)],
\end{equation}
where ${\lambda}_{kl}$ is the hyperparameter balancing the reconstruction loss. $p(z)$ is the prior and $q_\phi(z|x)$ is the posterior. 



    


\noindent \textbf{Auxiliary Supervised Guided-VAE.} 
The basic VAE described above maps a set of molecules $X$ into latent space and enables the sampling of valid molecules. However, such a model is not enforced to explicitly capture \emph{content} attributes. To enable effective optimization, the molecules encoded in the continuous representations of the VAE need to be correlated with the properties that we are seeking to retain. Inspired by \cite{ding2020guided}, we introduced an auxiliary guided-VAE to learn a transparent representation by introducing guidance to the latent variables as shown in Figure \ref{fig:fig2}.

Formally, for training data $X = (x_1, ..., x_m)$, suppose there exists a total of $T$ attributes with ground-truth labels. Let $z =(z_t, \mathbf{z}_{t}^{rst})$ where $z_t$ defines a scalar variable deciding the $t$-th attribute and $\mathbf{z}_{t}^{rst}$ represents the rest latent variables. For each attribute, we use an adversarial excitation and inhibition mechanism with term:
\begin{equation}
\&\mathcal{L}_{Excitation}(\phi,t)= \\
\&\max\limits_{{\nu_t}}   ~~\sum_{i=1}^{m}\mathbb{E}_{q_\phi(z_t|x_i)}[\log p_{\nu_t}(y=y_t(x_i)|z_t)]
\end{equation}
where $\nu_t$ denotes the parameters for an attribute predictor making a prediction for the $t$-th attribute through the latent variable $z_t$, and $y_t(x_i)$ is the ground-truth label for the $t$-th attribute of sample $x_i$. Specifically, the attribute predictor can be a binary classifier for existing of aromatic ring prediction or a regression model for molecular weight. This is an excitation process since we want latent variable $z_t$ to directly correspond to the attributes. Next is an inhibition term:
\begin{equation}
\&\mathcal{L}_{Inhibition}(\phi,t)= \\
\&\max\limits_{{\tau_t}} ~~{\sum_{i=1}^{m}\mathbb{E}_{q_\phi(\mathbf{z}_t^{rst}|x_i)}[\log p_{\tau_t}(y=y_t(x_i)|\mathbf{z}_t^{rst})] }
\end{equation}
where $\tau_t$ denotes the parameters for an inhibition predictor giving a prediction for the $t$-th attribute through the rest latent variables $\mathbf{z}_t^{rst}$. This is an inhibition term since the remaining variables $\mathbf{z}_t^{rst}$ is set independent from the attribute label $z_t$ in Eq. (\ref{eq:4}). The network is optimized by:
\begin{equation}\label{eq:4}
\&\max\limits_{{\theta,\phi}}~~{\sum_{i=1}^{m}\mathcal{L}_{KL}(\theta,\phi,x_i)+\\ \&\sum_{t=1}^{T}[\mathcal{L}_{Excitation}(\theta,\phi) - \mathcal{L}_{Inhibition}(\theta,\phi)] }
\end{equation}
Both the $\nu_t$ and $\tau_t$ are formed by two feed-forward layers and trained at the same time as the minimization of the VAE loss. Thus, this could be viewed as multi-task learning, incentivizing the model to not only decode the molecule but also predict its content from the latent vector. Similar idea has also been used in previous work for image \cite{ding2020guided} or sentence representation learning \cite{john2018disentangled}. 


\subsection{Learning Style Latent Space}

Previous style transfer works usually adopt VAE and single sample in style datasets to build style latent space. However, this approach is not expressive enough to express the stylistic feature as demonstrated in recent work \cite{kotovenko2019content, yi2020text}. Moreover, VAE assumes
the independence of molecules and endows each molecule with a corresponding isotropic Gaussian latent space, ignoring the inner connection of molecules with the same style. Taking inspiration from \cite{yi2020text}, we provide a set of molecules $\Phi_{K}^{j}=\{{m}^j_k\}_{k=1}^{K} \subset \mathcal{D}^j$, referred to \emph{molecular style instances}, to form the style latent space and further strengthen it with generative flow techniques \cite{rezende2015variational}.

Formally, let $z_0 \sim p(z_0|\Phi_{K}^{j})$ be an initial style latent variable,  we attempt to map $z_0$ to a more complicated one  $z_T$ by applying a sequence of $T$ transformation functions $f_t$:
\begin{equation}\label{eq5}
z_T= f_T(f_{T-1}(...f_1(z_{0},c)))
\end{equation}
where $T$ is the length of the sequence and $c$ is a conditional prior. According to \cite{rezende2015variational}, each $f_t$ should be invertible and its corresponding Jacobian determinant be easy to compute. To improve sampling efficiency, we adopt a simple but robust invertible function, Inverse Autoregressive Flow \cite{kingma2016improving} to accomplish this process:
\begin{gather}
[\varepsilon_t,\varphi_t]\leftarrow g_t(z_{t-1},c),
\sigma_t = \delta (\varphi_t)\label{eq7},\\
z_t = \sigma_t \odot z_{t-1}+(1-\sigma_t)\odot  \varepsilon_t,
\end{gather}
where $\delta$ is a $sigmoid$ activation function, $g_t$ is a masked autoregressive network proposed by \cite{germain2015made}, $\varepsilon_t$ and $\varphi_t$ are two output real-value vectors computed by $g_t$. Under the invertibility assumption for $f_t$, the final complicated probability density distribution $\pi(z_T|c)$ can be obtained by:

\begin{equation}
 \pi(z_T|c) = p(z_0|c) \prod_{t=1}^{T}   \left |\mathrm{det} \left(\frac {z_t} {z_{t-1}} \right) \right |
\end{equation}

Note that $c$ is a conditional prior derived from $\Phi_{K}^{j}$. To calculate $c$, we first assume the initial variable $z_0$ in Eq.(\ref{eq5}) follows the isotropic Gaussian distribution:
\begin{gather}
z_0 \sim p(z_0|\Phi_{K}^{j}) = N (\mu_0, \sigma_0^2 I)\label{eq9}, \\
\mu_0 \approx \frac{1}{K} \sum_{k=1}^{K}z_s,\sigma_0^2 \approx \frac{1}{K-1} \sum_{k=1}^{K}(z_s-\sigma_0^2)^2
\end{gather}
where the mean of $z_0$ is approximated by an unbiased estimator with the mean $\mu_0$ and error $\sigma_0^2$,  $z_s$ is computed by $Enc(m^j; \phi)$, where $m^j \in \Phi_{K}^{j}$.  $c$ is further computed by $MLP$ with two fully connected layers as $c= MLP(\mu_0)$. As such, we can obtain an expressive hidden style feature $h_s$ ($h_s$ = $z_T$) by sampling $z_0$ with Eq.(\ref{eq9}) and mapping it with Eq.(\ref{eq5}). Then the sampled $h_s$ is fed to the Generator $G$ with reference molecular latent representation $z_{c}$ to get the transferred latent representation $z_{g}$.

\begin{figure*}
	\centering
	\includegraphics[scale=0.62]{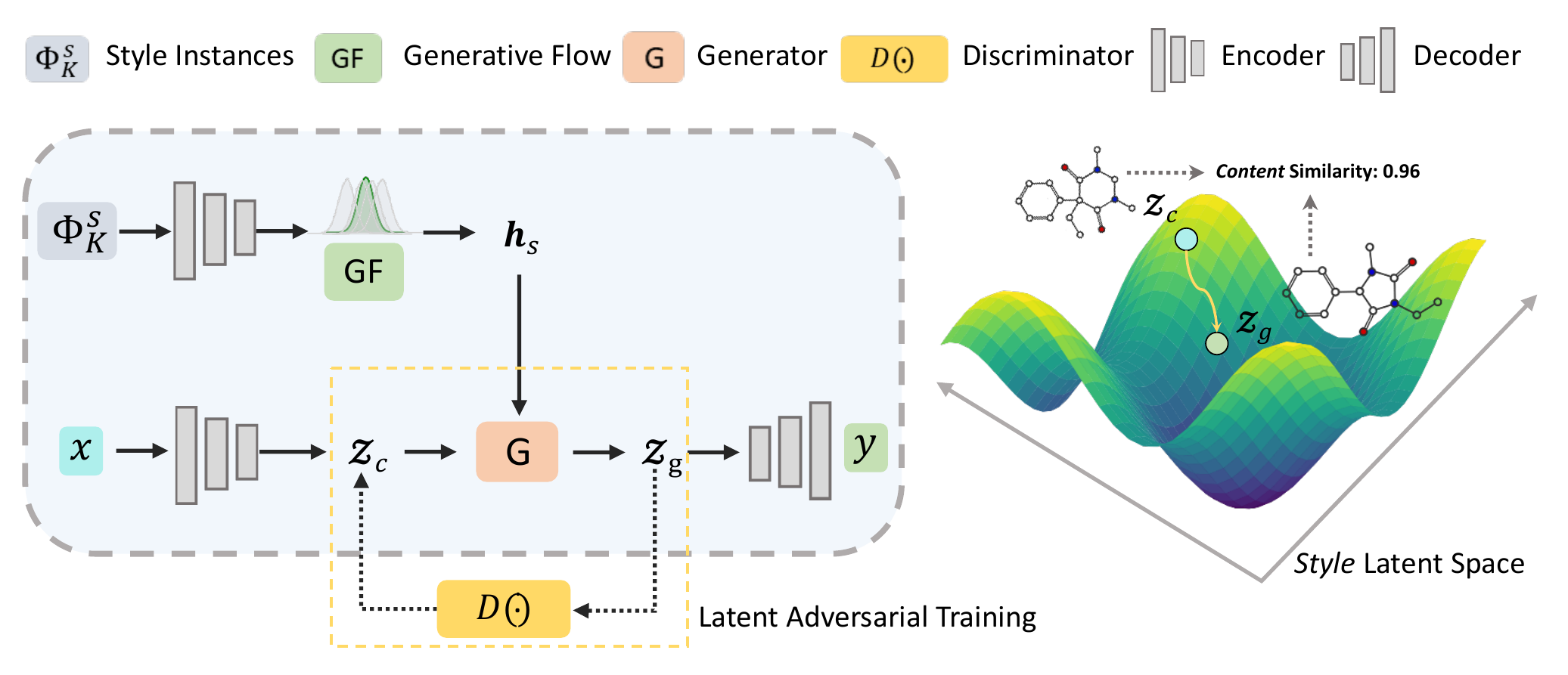}
	\vspace{-5mm}
	\caption{Pipeline overview.
	}
	\label{fig:overview}
\end{figure*}


\subsection{Adversarial Training and Auxiliary Losses}
Given a reference molecule $m^i$, two sets of molecular style instances, $\Phi_{K}^{i}$ ($m^i$ $\not\in$ $\Phi_{K}^{i}$) and $\Phi_{K}^{j}$, with the source style $i$ and the target style $j$, we adopt the latent adversarial training with two auxiliary loss functions to create circuitous supervision from non-parallel corpora.

\emph{Adversarial Style Loss.} Without any parallel corpus, adversarial
training \cite{goodfellow2014explaining} is utilized to control the attribute of the generated molecules. Different from the previous adversarial learning methods, we do not directly use SMILES strings as input, but instead latent vectors derived from the VAE encoder ($Enc(x; \phi)$) and Generator $G$. This allows the model to focus on optimizing the model in continuous latent space and not worry about the discrete issues. Similar to the idea of \cite{dai2019style}, we adopt a classifier with $S$+1 (here $S$ denotes style $i$ and $j$, $S$+1-$th$ class indicates a fake sample) classes as the discriminator $D$ to differentiate the style of an input molecule. Both the discriminator and generator in our model are formed by three feed-forward layers and trained following Wasserstein GAN \cite{gulrajani2017improved} with gradient penalty.

The generator $G$ is expected to confuse the discriminator by minimizing negative log-likelihood:
\begin{equation}
\mathcal{L}_{style}=-\mathrm{log}p_{D}(j|z_g)
,
\end{equation}
The discriminator is optimized alternately by minimizing negative log-likelihood of the corresponding class of style:
\begin{equation}
\mathcal{L}_{\mathrm{adv}}^S= -[\mathrm{log}p_{D}(S|z_c)+\mathrm{log}p_{D}(S|\hat{z_c})],
\end{equation}
\begin{equation}
\mathcal{L}_{\mathrm{adv}}^{S+1}= -\mathrm{log}p_{D}(S+1|z_g),
\end{equation}
where $\hat{z_c} \leftarrow G(z_c, GF(Enc(\Phi_{K}^{S}),\phi))$, $S$ could be $i$ or $j$. And the total adversarial loss is defined as:
\begin{equation}
\mathcal{L}_{\mathrm{adv}}= \mathcal{L}_{\mathrm{adv}}^i+\mathcal{L}_{\mathrm{adv}}^j+\mathcal{L}_{\mathrm{adv}}^{S+1},
\end{equation}

Furthermore, we introduce two auxiliary training losses to enhance the supervision of our model.

\emph{Self Reconstruction Loss.} The first loss function is used to reconstruct the latent vector $z_c$ of input molecules by minimizing negative log-likelihood:
\begin{equation}
\mathcal{L}_{recon}=-\mathrm{log}p_{G}(z_c|z_c,\Phi_{K}^i)
\end{equation}


\emph{Cycle Consistency Loss.}  To encourage generated molecule preserving the content of the input molecule, we feed the generated molecule latent representation $z_g$ back to our model with the $\Phi_{K}^{i}$ and train our model to reconstruct the original latent vector $z_c$ by minimizing negative log-likelihood:
\begin{equation}
\mathcal{L}_{cycle}=-\mathrm{log}p_{G}(z_c|z_g,\Phi_{K}^i)
\end{equation}


Overall, $\mathcal{L}_{adv}$ is designed for differentiating the style of the latent vector $z_c$ of input molecules; $\mathcal{L}_{recon}$ is utilized for reconstructing the latent vector $z_c$  given its source style signals, enabling the disentanglement of style and content. $\mathcal{L}_{cycle}$ is used to strengthen content preservation. All of these regularizers are performed over continuous latent space. The whole training process has been shown in Algorithm 1.





\begin{algorithm}[!htb]
\caption{Training Process}
\label{alg:algorithm}
\textbf{Input}: {Content dataset $\mathcal D^i$ with source style $i$ and style dataset $\mathcal D^j$ with target style $j$; ratio of discriminator to generator training frequency $N$;}

\textbf{Initialization}: Pretrain Auxiliary Guided-VAE
\begin{algorithmic}[1] 
\For{$k$ in number of iterations}
\State {Sample instances $\Phi_{K}^{j}$ from $\mathcal D{^j}$ and $\Phi_{K}^{i}$ from $\mathcal D{^i}$} 
\State Sample content $m{^i}$ from $\mathcal D{^i}$ 
\State Accumulate $\mathcal L{_{adv}}$
\If{$k\%N$ == 0}
\State Accumulate $\mathcal L{_{style}}, \mathcal L{_{recon}}, \mathcal L{_{cycle}}$ 
\State Update the parameters of $G, GF, D $
\Else

\State Update the parameters of $D$
\EndIf

\EndFor
\end{algorithmic}
\end{algorithm}



\section{Experiments}

\subsection{Datasets}
We conducted experiments on two molecular attributes transfer tasks.

\textbf{Toxicity modification}. Our first experiment focuses on molecular optimization with the goal of changing the underlying mutagenic toxicity, which can be regarded as style transfer from toxic to non-toxic molecules. The toxicity prediction model was trained on the dataset collected from \cite{hansen2009benchmark} with 3503 toxic compounds and 3009 non-toxic ones using graph neural networks \cite{yang2019analyzing}, which achieves an AUROC score of 89\%. We set the probability threshold of $p$ \textgreater 0.9 as toxic molecule and $p$ \textless 0.03 as non-toxic ones. Accordingly, the toxic compounds represent only 1.3\% of the ZINC lead-like dataset. We derived a set of 400K non-toxic molecule and 400K toxic ones from ZINC lead-like dataset, and respectively split into 85\%/5\%/10\% for train/dev/test.

\textbf{Synthesizability modification}. Our second set of experiments involves decreasing the molecular synthetic complexity, which has been a popular topic in organic chemistry. In particular, the model needs to decrease molecules with synthetic accessibility (SA) score \cite{ertl2009estimation} within the range [5,8] (hard to synthesize) into the lower range [0,2.5] (trivial to synthesize). This task is challenging as the SA may entangle with more basic physical properties. We also derived a set of 400K for trivial and hard molecules from ZINC lead-like dataset, respectively, and use an 85\%/5\%/10\% split for train/dev/test.




\subsection{Metrics}
An expected transferred molecule should be chemically valid, content-complete with the target style. To evaluate the performance of the different models,  we compared generated molecules from four different dimensions: 1) Style control, 2) Content preservation, 3) Validity, 4)Success rate.
\begin{itemize}[leftmargin=*]
\item \textbf{Style control}. We first predict the target property of generated molecules with the corresponding scoring functions. On the toxicity task, we use the learned toxicity predictor to measure mutagenic toxicity. On the SA task, we use the SA score to calculate the synthesizability of the molecules. We report the average Property Improvement \textbf{(Imp.)} of generated molecules in toxicity and synthesizability. It is defined as the difference of the property value (or probability) between generated molecules $Y$ and input molecule $X$, i.e., property($Y$) - property$X$).




\item \textbf{Content preservation}. To estimate the \textit{content} similarity between each reference compound and each generated sample from the initial pool in terms of MW, logP, number of HBA, number of HBD, and number of Rotate Bond, the number of ring, net charge and TPSA is calculated and summarized in a Physicochemical Similarity Score \textbf{(PSS)} \cite{vogel2011dekois}. This score consists of the arithmetic mean of the normalized similarity scores for each property. The PSS score is a real number in the range of (0,1], and the higher the value, the more similar the generated molecule is to the original input. The details of PSS are shown in appendix. Note that we did not use structural similarity as a metric, as all the chemical properties are determined by the molecular structure. 

\item \textbf{Validity} is the percentage of inferred SMILESs that are chemically valid according to RDKit \cite{landrum2013rdkit};

\item \textbf{Success Rate (SR)} To evaluate the overall transfer quality, we calculate success rate based on property improvement and content preservation (\emph{PSS}) between input molecule $X$ and generated molecule $Y$. In particular, we define a translation as successful if the translated compound simultaneously satisfies (a) predicted toxicity\textless 0.1 or SA\textless 2.5, and (b) the \emph{PSS} \textgreater 0.7.

\end{itemize}

Following the strategy in \cite{yi2020text}, we also report the geometric mean \textbf{(GM)} of Property Improvement, PSS and SR as an indication of the overall performance.



\subsection{Baselines} We compare our approach (MolStyle) with the following baselines: \textbf{1. \emph{Post-hoc}} filtering method. We performed \emph{post-hoc} filtering  with pre-defined screening constraints. For each reference molecule, we randomly matched 4K times in the ZINC lead-like library (equal to our test set size). The top molecule with the highest \emph{PSS} is selected as the final candidate. This is a strong baseline allowing us to find \emph{content} similar molecules. \textbf{2. REINVENT} \cite{olivecrona2017molecular} is a SMILES-based RL model for single-objective molecule generation. To generate realistic molecules, we used the original pre-trained model and then fine-tuned under target \emph{style} rewards. \textbf{3. GCPN} \cite{you2018graph} is a graph-based RL model that modifies a molecule by iteratively adding or deleting atoms and bonds. They also adopt adversarial training to enforce validity of the generated molecules. we re-trained the model under target \emph{style} rewards. \textbf{4. RationaleRL} \cite{jin2020multi} learns a graph completion model, but relies on a fixed set of multi-property rationales composed by single-property rationales extracted by MCTS. We set the thresholds of 0.5$\textless$ for toxicity and 2.5$\textless$ for synthesizability to extract the rationales and then complete the graph with the target style scoring function.  \textbf{5. StyIns}  \cite{yi2020text} is a state-of-the-art text style transfer method that adopts the attention mechanism to preserve source information and learns to extract stylistic properties from multiple instances. We followed the settings in the original paper but changed the inputs from sentences to SMILESs. More implementation details have been shown in the Appendix.

\subsection{Implementation details.} All molecules are represented in canonical SMILES, which are tokenized with the regular expression in \cite{olivecrona2017molecular}. We use Pytorch \cite{paszke2019pytorch}  to implement our models. We train the auxiliary guided-VAE on ZINC lead-like database for 80 epochs, optimizing with Adam \cite{kingma2014adam} with default settings. All the \emph{content} attribute predictor is formed by two fully feed-forward layers. It takes about 30 hours on two GTX 2080 Ti GPUs. Finally, it is able to generate 98.6\% valid molecules. We do not fine-tune the VAE after pre-training. We set token embedding size, VAE hidden state size,  the number of molecular style instances $K$, and the length of generative flow chain $T$ to 512, 512, 10, and 6, respectively. During inference, we follow the similar evaluation protocol as \cite{jin2018learning}. For each source molecule, we decode $k$ times with different latent variables $z_g$ $\sim$ $N (0, I)$, and report the molecule having the highest property improvement under the PSS constraint. We also applied the same sampling strategy for the other baseline models and set $k$=10. 




\begin{table*}
    \centering
      \caption{ \label{t1}\textbf{ Comparison of the effectiveness of attributes transfer tasks.} The best performance is highlighted. Both StyIns and MolStyle are unsupervised methods.}
    \resizebox{1\columnwidth}{!}
    {
        \begin{tabular}{c|c|c|c|c|c||c|c|c|c|c}
        
             \hline
              & \multicolumn{5}{c||}{Toxicity} & \multicolumn{5}{c}{Synthesizability}  \\
             \hline
             Methods  & Imp.$\uparrow$  & PSS $\uparrow$ & Val. $\uparrow$ & SR $\uparrow$  & GM $\uparrow$ & Imp. $\uparrow$  & PSS $\uparrow$ & Val. $\uparrow$ & SR $\uparrow$ & GM $\uparrow$ \\
             \hline
              ZINC (Random) & 0.723 & 0.727 & -- & 0.150 & 0.429 & 2.366 & 0.698 & -- & 0.162 & 0.644  \\
             ZINC (\emph{Post-hoc}) & 0.711 & \underline{0.859} & -- & \underline{0.216} & \underline{0.509} & 2.036 & \textbf{0.836} & -- & 0.161 & 0.653  \\
             \hline
             REINVENT  & \underline{0.857} &  0.607 &  93.2 &  0.196 & 0.467 & 3.197 &  0.609 &  91.6 &  \underline{0.264} &  0.809 \\
             GCPN  &  0.666 & 0.655 & \textbf{100} & 0.085 & 0.334 & 0.243 & 0.653 & \textbf{100} & 0.007 & 0.181  \\
             RationaleRL  & 0.843 & 0.657 & \textbf{100} & 0.181 & 0.464 & 2.962 & 0.692 & \textbf{100} & 0.216 & 0.762 \\
            \hline
            StyIns  & 0.332 & \textbf{0.895} & 35.8 & 0.073 & 0.279 & \underline{3.418} & 0.708 & 34.4 & 0.340 & \underline{0.937} \\
            MolStyle & \textbf{0.871} & 0.789 & \underline{98.8} & \textbf{0.589} & \textbf{0.739} & 3.068 & \underline{0.741} & \underline{98.2} & \textbf{0.472} & \textbf{1.024} \\
            \hline
        \end{tabular}
    }
\end{table*}%

\begin{wrapfigure}{r}{0.5\textwidth}
\centering
\includegraphics[width=0.5\textwidth]{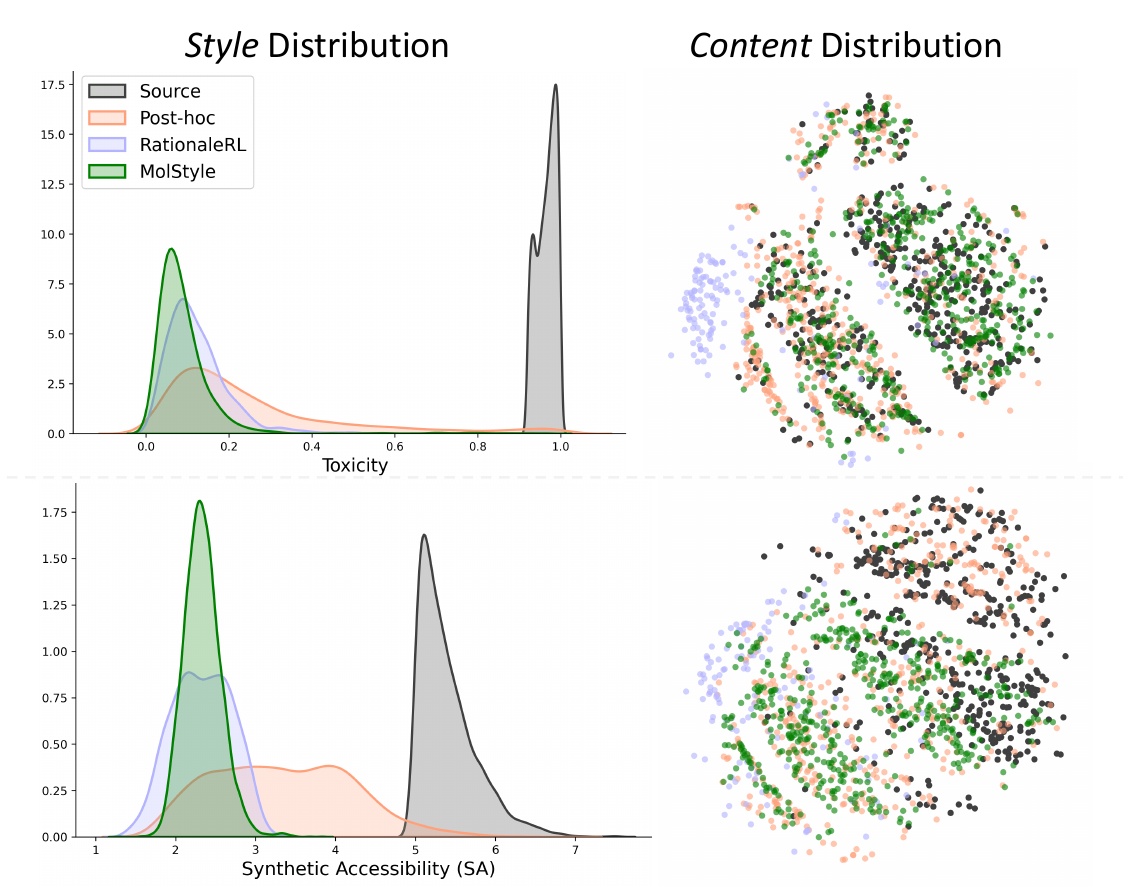}
\vspace{-7mm}
\caption{Visualization of \emph{style} and \emph{content} distributions of molecular samples generated from the different methods on two datasets. }
\label{fig:Model}
\end{wrapfigure}

\subsection{Experiments results}

Table 1 summarizes the attribute transfer performance of each model, and random compounds matched in ZINC lead-like dataset are also shown for comparison (\emph{ZINC random}). \emph{Post-hoc} filtering method achieves higher content preservation but lower style transfer accuracy and success rate compared to deep learning based-generation methods, demonstrating the limitation of the similarity-based searching model. Among RL-based generative models, single-objective models (e.g., REINVENT and GCPN) get satisfactory improvements of target property on both the Toxicity and Synthesizability datasets, but perform worse in PSS for content preservation. This is because  these methods are lacking  anchor molecules as constraints, and generate biased molecules with desired property but irrelevant content. Though RationaleRL is better at content preservation because of its powerful rationale extraction module, it doesn't achieve high attribute transfer accuracy. This is mainly because a simple modification on original molecular structure may cause huge perturbations in molecular properties, which is also known as \emph{acitivity cliff}. StyIns performs pretty well in synthesizability but the worst in toxicity. This suggests that StyIns can capture simple molecular attributes but have difficulties in complex chemical properties. The lowest valid rate is also in line with our expectation that discrete adversarial regularization will cause SMILES syntax issues. Generally, our MolStyle model, even without the help of scoring functions, achieves the best overall performance consistently (SR and GM).



In Figure 4, we vectorized generated molecules with their eight \emph{content} properties and projected them into chemical space by t-SNE with default hyper-parameters. We also visualized the corresponding \emph{style} properties distribution. We can observe that RationaleRL achieves a high style transfer accuracy but low degree of content overlap with reference molecules while \emph{post-hoc} is almost the opposite. By contrast, our model achieve both high style transfer accuracy and high content preservation, indicating the good disentanglement capability of our model. We note that the relatively worse content overlap of SA is expected, as SA is proved to be highly correlated with several basic properties (e.g., logP and TPSA) \cite{ertl2009estimation}. Several cases sampled by our model has been shown in Figure 5.

\subsection{Further analysis} 
\noindent \textbf{Ablation study} We conducted ablation studies on the Toxicity dataset to investigate factors that influence the performance of the proposed MolStyle framework. As shown in Table 2, MolStyle with all the essential modules delivers the best performance among all architectures. The exclusion of all the key modules performed the worst. The exclusions of the generative flow and cycle consistency loss both caused decreases in performances. The use of Auxiliary Guidance is beneficial by increasing the GM from 0.593 to 0.739, indicating the importance of constructing a discriminative VAE model.


\begin{figure*}
	\centering
	\includegraphics[width=1\textwidth]{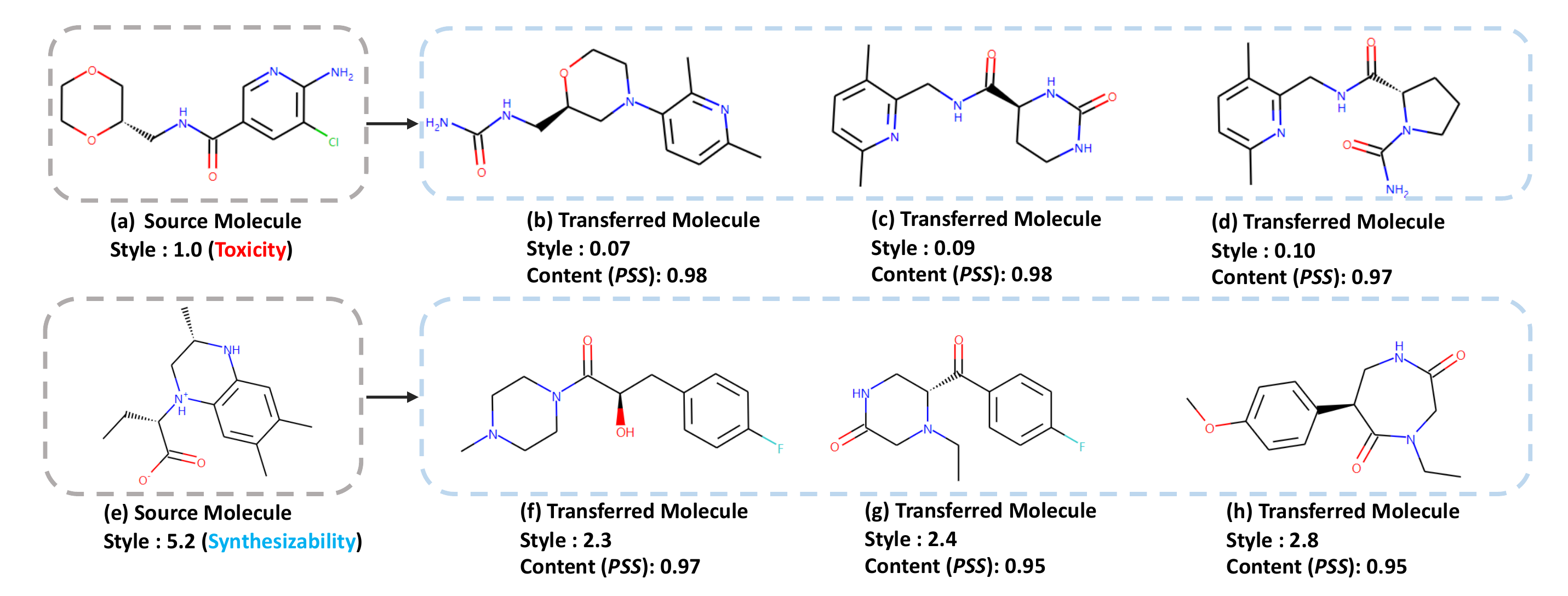}
	\vspace{-7mm}
	\caption{Examples of Toxicity (Upper) and Synthesizability (Lower) optimization. We report the target attribute value (Style) and the physicochemical similarity score (PSS) between generated molecules and input ones. 
	}
	\label{fig:overview}
\end{figure*}

\noindent \textbf{Toxicity analysis study}  As the reported performance is mostly based on learned property predictors, it is possible for the model to exploits the loopholes of the predictors. As a result, the predicted properties may be unreliable when the distribution of generated compounds is very different from the distribution of molecules used to train the property predictors. To enable a trustworthy evaluation, we furhter adopt the structural alert-based metric to measure the effectiveness of our model. In particular, we collected 46 toxicophores that have been proved to be the key components of mutagenicity \cite{kazius2005derivation}. Each molecule is labeled as toxic if it contains at least one toxicophores (i.e., substructures) in the database. Then we counted how many of the generated molecules containing these alerts. We compare our model with RationaleRL and REINVENT models. As shown in Table 3, our method outperforms the baselines in terms of the structural alert mapping, with only 18.8\% molecules match on average, improving the RationaleRL and Reinvent by 2.9\% and 11.8\%. This is expected as the rationale-based methods try to modify the molecular structures based on the substructure of reference molecules, which probably maintain some toxic alerts. In contrast, our model is capable of removing the unfavorable substructures while retaining the key content embedded in the input molecules.

\begin{table*}
  \begin{minipage}{0.48\textwidth}
\centering
\resizebox{.95\columnwidth}{!}{\begin{tabular}{l|c c c c}
 \hline
Metrics & Imp. & PSS & SR & GM \\
\hline
w/o All & 0.862 & 0.699 &  0.297&  0.564  \\
w/o $\mathcal{L}_{recon}$ & 0.875 & 0.741 &  0.463 & 0.670 \\
w/o $GF$ & 0.859 & {\bf0.793} &  0.482 &  0.690  \\
w/o Aux\_Guidance & {\bf0.877} & 0.751 & 0.316 & 0.593 \\
\hline
MolStyle  & 0.871 & 0.789 & \textbf{0.589} & \textbf{0.739}\\
\hline
\end{tabular}}
\caption{Ablation results on Toxicity datasets. The removal of the basic modules decreased the overall performance of MolStyle.}
\label{tab: Ablation results}
\end{minipage}
\hfill
\begin{minipage}{0.5\textwidth}
\resizebox{.87\columnwidth}{!}{\begin{tabular}{c| c c c c}
\hline
Metrics & Alert$\downarrow$ & Imp.$\uparrow$ & PSS$\uparrow$  & SR$\uparrow$ \\
\hline
Source & 26.1\% & -- &  -- \\

REINVENT & 30.6\% & 0.857 &  0.607  &  0.196\\
RationaleRL & 21.7\% & 0.843 &  0.657 &  0.181\\
\hline
MolStyle & {\bf 18.8\%} & {\bf 0.871} &  {\bf 0.789}  &  {\bf0.589}\\
\hline
\end{tabular}}
\caption{Structural alert-based evaluation on the toxicity dataset. Our model is capable of removing the unfarvorable substructures while retaining the key content embedded in the input molecules.}
\label{tab: Ablation results}
\end{minipage}
\end{table*}




\section{Conclusions}
In this paper, for the first time, we formulate molecular optimization as a style transfer problem and present a novel training algorithm that could automatically learn internal differences between two groups of non-parallel data through adversarial training strategies. Experiments on two new molecular optimization tasks demonstrate that our model significantly outperforms several state-of-the-art methods. 


\begin{ack}

We would like to thank Binghong Chen and Hanjun Dai for discussion.
This study has been supported by the National Key R\&D Program of China
(2020YFB020003), National Natural Science Foundation of China (61772566),
Guangdong Key Field R\&D Plan (2019B020228001 and 2018B010109006),
Introducing Innovative and Entrepreneurial Teams (2016ZT06D211), Guangzhou S\&T Research Plan (202007030010).

\end{ack}

{\footnotesize
\bibliographystyle{unsrt}  
\bibliography{bibfile} 

\begin{thebibliography}{10}

\bibitem{nicolaou2007molecular}
Christos~A Nicolaou, Nathan Brown, and Constantinos~S Pattichis.
\newblock Molecular optimization using computational multi-objective methods.
\newblock {\em Current Opinion in Drug Discovery and Development}, 10(3):316,
  2007.

\bibitem{polishchuk2013estimation}
Pavel~G Polishchuk, Timur~I Madzhidov, and Alexandre Varnek.
\newblock Estimation of the size of drug-like chemical space based on gdb-17
  data.
\newblock {\em Journal of computer-aided molecular design}, 27(8):675--679,
  2013.

\bibitem{blaschke2020reinvent}
Thomas Blaschke, Josep Ar{\'u}s-Pous, Hongming Chen, Christian Margreitter,
  Christian Tyrchan, Ola Engkvist, Kostas Papadopoulos, and Atanas Patronov.
\newblock Reinvent 2.0: an ai tool for de novo drug design.
\newblock {\em Journal of Chemical Information and Modeling},
  60(12):5918--5922, 2020.

\bibitem{you2018graph}
Jiaxuan You, Bowen Liu, Rex Ying, Vijay Pande, and Jure Leskovec.
\newblock Graph convolutional policy network for goal-directed molecular graph
  generation.
\newblock {\em arXiv preprint arXiv:1806.02473}, 2018.

\bibitem{lin1997role}
Jiunn~H Lin and Anthony~YH Lu.
\newblock Role of pharmacokinetics and metabolism in drug discovery and
  development.
\newblock {\em Pharmacological reviews}, 49(4):403--449, 1997.

\bibitem{li2018multi}
Yibo Li, Liangren Zhang, and Zhenming Liu.
\newblock Multi-objective de novo drug design with conditional graph generative
  model.
\newblock {\em Journal of cheminformatics}, 10(1):1--24, 2018.

\bibitem{winter2019efficient}
Robin Winter, Floriane Montanari, Andreas Steffen, Hans Briem, Frank No{\'e},
  and Djork-Arn{\'e} Clevert.
\newblock Efficient multi-objective molecular optimization in a continuous
  latent space.
\newblock {\em Chemical science}, 10(34):8016--8024, 2019.

\bibitem{jin2020multi}
Wengong Jin, Regina Barzilay, and Tommi Jaakkola.
\newblock Multi-objective molecule generation using interpretable
  substructures.
\newblock In {\em International Conference on Machine Learning}, pages
  4849--4859. PMLR, 2020.

\bibitem{jin2018learning}
Wengong Jin, Kevin Yang, Regina Barzilay, and Tommi Jaakkola.
\newblock Learning multimodal graph-to-graph translation for molecular
  optimization.
\newblock {\em arXiv preprint arXiv:1812.01070}, 2018.

\bibitem{maggiora2014molecular}
Gerald Maggiora, Martin Vogt, Dagmar Stumpfe, and Jurgen Bajorath.
\newblock Molecular similarity in medicinal chemistry: miniperspective.
\newblock {\em Journal of medicinal chemistry}, 57(8):3186--3204, 2014.

\bibitem{weininger1988smiles}
David Weininger.
\newblock Smiles, a chemical language and information system. 1. introduction
  to methodology and encoding rules.
\newblock {\em Journal of chemical information and computer sciences},
  28(1):31--36, 1988.

\bibitem{gomez2018automatic}
Rafael G{\'o}mez-Bombarelli, Jennifer~N Wei, David Duvenaud, Jos{\'e}~Miguel
  Hern{\'a}ndez-Lobato, Benjam{\'\i}n S{\'a}nchez-Lengeling, Dennis Sheberla,
  Jorge Aguilera-Iparraguirre, Timothy~D Hirzel, Ryan~P Adams, and Al{\'a}n
  Aspuru-Guzik.
\newblock Automatic chemical design using a data-driven continuous
  representation of molecules.
\newblock {\em ACS central science}, 4(2):268--276, 2018.

\bibitem{olivecrona2017molecular}
Marcus Olivecrona, Thomas Blaschke, Ola Engkvist, and Hongming Chen.
\newblock Molecular de-novo design through deep reinforcement learning.
\newblock {\em Journal of cheminformatics}, 9(1):1--14, 2017.

\bibitem{jin2018junction}
Wengong Jin, Regina Barzilay, and Tommi Jaakkola.
\newblock Junction tree variational autoencoder for molecular graph generation.
\newblock In {\em International Conference on Machine Learning}, pages
  2323--2332. PMLR, 2018.

\bibitem{zheng2019qbmg}
Shuangjia Zheng, Xin Yan, Qiong Gu, Yuedong Yang, Yunfei Du, Yutong Lu, and Jun
  Xu.
\newblock Qbmg: quasi-biogenic molecule generator with deep recurrent neural
  network.
\newblock {\em Journal of cheminformatics}, 11(1):1--12, 2019.

\bibitem{lim2020scaffold}
Jaechang Lim, Sang-Yeon Hwang, Seokhyun Moon, Seungsu Kim, and Woo~Youn Kim.
\newblock Scaffold-based molecular design with a graph generative model.
\newblock {\em Chemical Science}, 11(4):1153--1164, 2020.

\bibitem{fu2020core}
Tianfan Fu, Cao Xiao, and Jimeng Sun.
\newblock Core: Automatic molecule optimization using copy \& refine strategy.
\newblock In {\em Proceedings of the AAAI Conference on Artificial
  Intelligence}, volume~34, pages 638--645, 2020.

\bibitem{jin2019hierarchical}
Wengong Jin, Regina Barzilay, and Tommi Jaakkola.
\newblock Hierarchical graph-to-graph translation for molecules.
\newblock {\em arXiv preprint arXiv:1907.11223}, 2019.

\bibitem{zheng2020deep}
Shuangjia Zheng, Zengrong Lei, Haitao Ai, Hongming Chen, Daiguo Deng, and
  Yuedong Yang.
\newblock Deep scaffold hopping with multi-modal transformer neural networks.
\newblock 2020.

\bibitem{shen2017style}
Tianxiao Shen, Tao Lei, Regina Barzilay, and Tommi Jaakkola.
\newblock Style transfer from non-parallel text by cross-alignment.
\newblock {\em arXiv preprint arXiv:1705.09655}, 2017.

\bibitem{hu2017toward}
Zhiting Hu, Zichao Yang, Xiaodan Liang, Ruslan Salakhutdinov, and Eric~P Xing.
\newblock Toward controlled generation of text.
\newblock In {\em International Conference on Machine Learning}, pages
  1587--1596. PMLR, 2017.

\bibitem{lample2018multiple}
Guillaume Lample, Sandeep Subramanian, Eric Smith, Ludovic Denoyer,
  Marc'Aurelio Ranzato, and Y-Lan Boureau.
\newblock Multiple-attribute text rewriting.
\newblock In {\em International Conference on Learning Representations}, 2018.

\bibitem{john2018disentangled}
Vineet John, Lili Mou, Hareesh Bahuleyan, and Olga Vechtomova.
\newblock Disentangled representation learning for non-parallel text style
  transfer.
\newblock {\em arXiv preprint arXiv:1808.04339}, 2018.

\bibitem{shi2018toward}
Zhan Shi, Xinchi Chen, Xipeng Qiu, and Xuanjing Huang.
\newblock Toward diverse text generation with inverse reinforcement learning.
\newblock {\em arXiv preprint arXiv:1804.11258}, 2018.

\bibitem{sanchez2017optimizing}
Benjamin Sanchez-Lengeling, Carlos Outeiral, Gabriel~L Guimaraes, and Alan
  Aspuru-Guzik.
\newblock Optimizing distributions over molecular space. an
  objective-reinforced generative adversarial network for inverse-design
  chemistry (organic).
\newblock {\em ChemRxiv}, 2017, 2017.

\bibitem{dai2018syntax}
Hanjun Dai, Yingtao Tian, Bo~Dai, Steven Skiena, and Le~Song.
\newblock Syntax-directed variational autoencoder for structured data.
\newblock {\em arXiv preprint arXiv:1802.08786}, 2018.

\bibitem{faulon2003signature}
Jean-Loup Faulon, Donald~P Visco, and Ramdas~S Pophale.
\newblock The signature molecular descriptor. 1. using extended valence
  sequences in qsar and qspr studies.
\newblock {\em Journal of chemical information and computer sciences},
  43(3):707--720, 2003.

\bibitem{kingma2013auto}
Diederik~P Kingma and Max Welling.
\newblock Auto-encoding variational bayes.
\newblock {\em arXiv preprint arXiv:1312.6114}, 2013.

\bibitem{chung2014empirical}
Junyoung Chung, Caglar Gulcehre, KyungHyun Cho, and Yoshua Bengio.
\newblock Empirical evaluation of gated recurrent neural networks on sequence
  modeling.
\newblock {\em arXiv preprint arXiv:1412.3555}, 2014.

\bibitem{ding2020guided}
Zheng Ding, Yifan Xu, Weijian Xu, Gaurav Parmar, Yang Yang, Max Welling, and
  Zhuowen Tu.
\newblock Guided variational autoencoder for disentanglement learning.
\newblock In {\em Proceedings of the IEEE/CVF Conference on Computer Vision and
  Pattern Recognition}, pages 7920--7929, 2020.

\bibitem{kotovenko2019content}
Dmytro Kotovenko, Artsiom Sanakoyeu, Sabine Lang, and Bjorn Ommer.
\newblock Content and style disentanglement for artistic style transfer.
\newblock In {\em Proceedings of the IEEE/CVF International Conference on
  Computer Vision}, pages 4422--4431, 2019.

\bibitem{yi2020text}
Xiaoyuan Yi, Zhenghao Liu, Wenhao Li, and Maosong Sun.
\newblock Text style transfer via learning style instance supported latent
  space.
\newblock IJCAI, 2020.

\bibitem{rezende2015variational}
Danilo Rezende and Shakir Mohamed.
\newblock Variational inference with normalizing flows.
\newblock In {\em International conference on machine learning}, pages
  1530--1538. PMLR, 2015.

\bibitem{kingma2016improving}
Diederik~P Kingma, Tim Salimans, Rafal Jozefowicz, Xi~Chen, Ilya Sutskever, and
  Max Welling.
\newblock Improving variational inference with inverse autoregressive flow.
\newblock {\em arXiv preprint arXiv:1606.04934}, 2016.

\bibitem{germain2015made}
Mathieu Germain, Karol Gregor, Iain Murray, and Hugo Larochelle.
\newblock Made: Masked autoencoder for distribution estimation.
\newblock In {\em International Conference on Machine Learning}, pages
  881--889. PMLR, 2015.

\bibitem{goodfellow2014explaining}
Ian~J Goodfellow, Jonathon Shlens, and Christian Szegedy.
\newblock Explaining and harnessing adversarial examples.
\newblock {\em arXiv preprint arXiv:1412.6572}, 2014.

\bibitem{dai2019style}
Ning Dai, Jianze Liang, Xipeng Qiu, and Xuanjing Huang.
\newblock Style transformer: Unpaired text style transfer without disentangled
  latent representation.
\newblock {\em arXiv preprint arXiv:1905.05621}, 2019.

\bibitem{gulrajani2017improved}
Ishaan Gulrajani, Faruk Ahmed, Martin Arjovsky, Vincent Dumoulin, and Aaron
  Courville.
\newblock Improved training of wasserstein gans.
\newblock {\em arXiv preprint arXiv:1704.00028}, 2017.

\bibitem{hansen2009benchmark}
Katja Hansen, Sebastian Mika, Timon Schroeter, Andreas Sutter, Antonius
  Ter~Laak, Thomas Steger-Hartmann, Nikolaus Heinrich, and Klaus-Robert Muller.
\newblock Benchmark data set for in silico prediction of ames mutagenicity.
\newblock {\em Journal of chemical information and modeling}, 49(9):2077--2081,
  2009.

\bibitem{yang2019analyzing}
Kevin Yang, Kyle Swanson, Wengong Jin, Connor Coley, Philipp Eiden, Hua Gao,
  Angel Guzman-Perez, Timothy Hopper, Brian Kelley, Miriam Mathea, et~al.
\newblock Analyzing learned molecular representations for property prediction.
\newblock {\em Journal of chemical information and modeling}, 59(8):3370--3388,
  2019.

\bibitem{ertl2009estimation}
Peter Ertl and Ansgar Schuffenhauer.
\newblock Estimation of synthetic accessibility score of drug-like molecules
  based on molecular complexity and fragment contributions.
\newblock {\em Journal of cheminformatics}, 1(1):1--11, 2009.

\bibitem{vogel2011dekois}
Simon~M Vogel, Matthias~R Bauer, and Frank~M Boeckler.
\newblock Dekois: Demanding evaluation kits for objective in silico screening a
  versatile tool for benchmarking docking programs and scoring functions.
\newblock {\em Journal of chemical information and modeling},
  51(10):2650--2665, 2011.

\bibitem{landrum2013rdkit}
Greg Landrum.
\newblock Rdkit documentation.
\newblock {\em Release}, 1(1-79):4, 2013.

\bibitem{paszke2019pytorch}
Adam Paszke, Sam Gross, Francisco Massa, Adam Lerer, James Bradbury, Gregory
  Chanan, Trevor Killeen, Zeming Lin, Natalia Gimelshein, Luca Antiga, et~al.
\newblock Pytorch: An imperative style, high-performance deep learning library.
\newblock {\em arXiv preprint arXiv:1912.01703}, 2019.

\bibitem{kingma2014adam}
Diederik~P Kingma and Jimmy Ba.
\newblock Adam: A method for stochastic optimization.
\newblock {\em arXiv preprint arXiv:1412.6980}, 2014.

\bibitem{kazius2005derivation}
Jeroen Kazius, Ross McGuire, and Roberta Bursi.
\newblock Derivation and validation of toxicophores for mutagenicity
  prediction.
\newblock {\em Journal of medicinal chemistry}, 48(1):312--320, 2005.

\end{thebibliography}
}

\end{document}